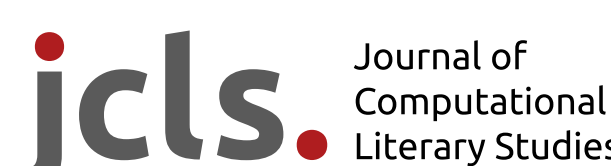

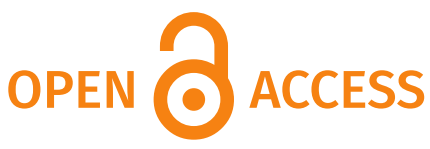

# Modeling Narrative Structure in Latin Epic Poetry with Automatically Generated Story Grammars

Abigail Swenor[1]
John James[2]
Neil Coffee[2]
Walter Scheirer[1]

1. Department of Computer Science and Engineering, University of Notre Dame, South Bend, USA.
2. Department of Classics, University at Buffalo, Buffalo, USA.

**Abstract.** Computational methods for analyzing prose and poetry utilize word embeddings and other abstract representations that sometimes obscure context-rich literary text. Inspired by the psychology of reading, we utilize story structure and elements to simulate human narrative comprehension to produce a more comprehensive representation of literary text. We present a method for automatically generating story grammar labels for input texts as a means of analysis that is interpretable and accessible by humanists and technologists alike. Using a large language model (LLM) pipeline and few-shot learning, we label Latin epic poetry with story element labels and use this output directly to aid an analysis of the story structure and style. Our method guides literary scholars to discover new areas of interest across texts and provides a new feature set for further study for downstream machine learning tasks.

## 1. Introduction

Over the past couple of decades, distant reading (Moretti 2005) has matured as a quantitative literary analysis approach that is distinct from traditional close reading. To perform distant reading, a computational method is applied to the entirety of a literary corpus, abstracting information about the text in order to make the analysis tractable for scholars, and ideally surfacing patterns that would otherwise not be recognizable in a close reading. Traditional methods for distant reading are specialized for specific literary analysis tasks (Bolt et al. 2025; Bothwell et al. 2023; Chaudhuri et al. 2015; Chaudhuri and Dexter 2017; Coffee et al. 2012; Forstall et al. 2015; Sprugnoli et al. 2023) and are therefore limited in their application. However, this is not a flaw but rather done by design. To stay true to the process of scholarship, it is important to understand the context in which one is working. For example, for a close reading, one would not want to use the same techniques for analyzing Latin epic poetry and contemporary American fiction. Thus, most distant reading approaches contain some level of specificity for a particular textual domain. To avoid relying on tailored datasets and models for each text and domain for which we want to apply distant reading and computational methods, we introduce story grammars to provide a human-readable and generalizable feature set.

Current distant reading approaches tend to utilize machine learning models that transform text to vector space embeddings — from semantically meaningful text to rows of numbers representing abstract features of that text. While this is useful for the machine learning models — which are then trained to complete the specific task at hand using this representation — it limits the interpretability and generalizability of the features that are learned through a model's training process. The features are narrowly tailored to the task on which the model was trained, and an examination of them at the embedding level would reveal little in the way of literary richness.

In this paper, we aim to create a human-readable feature set that can be abstracted from a wide variety of literary texts. Taking inspiration from research on the psychology of reading, we developed our approach by first studying the tools that readers use for narrative comprehension. In the last several decades, psychologists have concluded that readers understand text through the implicit use of story grammars (Frisch and Perlis 1981; Olson and Gee 1988; Taylor 1983; Zahoor and Janjua 2013). As shown in those previous studies, analyzing narratives according to story grammar elements closely follows real human interaction with a story. It is one major way that we make sense of, follow, compare, and compose narratives. In order to replicate this human interaction with the text computationally, our method specifically uses linguistic methods, such as transformational grammars (Chomsky 1956) that have a set of rules to process natural language according to defined transformations. The result of these transformations is a passage with a set of story element labels. In our study, we use language modeling to label literary texts with story elements, which we use as a tool for novel literary analysis for Latin epic poetry. We showcase how these story grammar labels can be applied to various literary analysis tasks — such as, semantic matching, discovering story structure patterns, and stylometric analysis — and demonstrate the validity of these labels by examining two Latin poets through a comparison of their use of story elements and narrative structures.

## 2. Related Work

Psychologists who study the process of reading recognize that there are standard elements of a story such as characters, plot, themes, and setting. These story elements can become more complex as the stories being analyzed become more complex. However, each narrative can be broken down into these basic story elements. Story grammars are a way of organizing and specifying relations between story elements. This structural approach to narrative also has important antecedents in Russian Formalism, especially in Vladimir Propp's analysis of folktales. Propp demonstrated that narratives could be decomposed into a finite set of recurring functions arranged in constrained sequences, independent of their specific lexical or thematic realization (Propp 2010). His work anticipates later story-grammar models by treating narrative structure as an abstract system underlying surface variation.

Our adaptation of these story grammars is formalized in the same way as Chomsky's transformational grammars and phrase structure (Chomsky 1956), in which story grammars should be able to generate all possible legitimate stories and no non-stories (Taylor 1983). Taylor 1983 outlined the history of using story grammars to model reading comprehension and provided multiple examples of possible story grammars. The use of this reading comprehension tool has been studied (Olson and Gee 1988; Zahoor and Janjua 2013) and even questioned over time (Frisch and Perlis 1981). The application of formal language theory in the original conception of story grammars was methodologically confusing, so Frisch and Perlis 1981 suggested a new formalization and emphasize that story grammars be evaluated by the knowledge they embody and not by the language which they use to express that knowledge. Therefore, story grammars remain a helpful tool in mapping structure and understanding reading comprehension.

Story grammars function in a similar way to transformational grammars and the same terminology can be used to describe them. Chomsky believed that each sentence has a surface structure and a deep structure. A deep structure is the mental model in a reader's mind. The rules in a grammar transform the deep structure into the surface structure which corresponds to the actual words read on the page (Pinker 1998)[1]. In literature, the deep structure could be seen as the semantic meaning of a given sentence while the surface structure is the exact words and phrases that make up the sentence. Each grammar is made up of transformational rules which can have multiple layers of structure that allow for multiple paths, or strings, to be created. For example, a rule could have multiple options for transformation: $START \longrightarrow LEFT|RIGHT$ where there is an option to choose the $LEFT$ or $RIGHT$ transformation. All three of these tokens, $START$, $LEFT$, and $RIGHT$, are examples of non-terminals. Non-terminals are placeholders for a set of terminals that can be generated by the transformation of non-terminals. A start symbol, such as $START$, is a special non-terminal where the grammar must begin with the first transformational rule. This terminology and structure is the same as seen in story grammars (Franzosi 2010; Taylor 1983).

In the study of classics — and more particularly digital classics — computational methods are a common way to perform distant reading and other associated downstream analyses (Bolt et al. 2025), such as semantic matching and allusion detection (Chaudhuri et al. 2015; Chaudhuri and Dexter 2017; Forstall et al. 2015), rhetorical parallelism (Bothwell et al. 2023), and sentiment analysis (Sprugnoli et al. 2023). All of these previous works in digital classics vary in their methodologies, but they have something in common: they require computational methods to assist with literary analysis of classical prose and poetry. Similarly, our work investigates the use of Large Language Models (LLMs) to provide a new way of approaching distant reading in Latin epic poetry. Previous studies have tackled established tasks in both natural language processing (NLP) and classical studies, but on a case-by-case basis. Our method aims to adapt the long-standing story grammar concept for computational methods to provide an interpretable and generalizable feature set that can be applied to a wide variety of these established literary analysis tasks.

## 3. Method

Our method outlined in this section uses a GPT-5 model (OpenAI 2025) that produces labels for Latin text drawn from our story grammar structure. We performed few-shot

1. While contemporary linguistic theory has largely moved beyond classic transformational grammars, its core abstractions remain useful as a conceptual model for how story grammars operate — namely, the distinction between underlying structural representations and their surface realizations.

learning to teach the model how to assign the labels in the structure we intended. This method for training is used when labeled data is scarce and it relies on the ability of machine learning models to learn with access to a small amount of training data (Wang et al. 2020b). In few-shot learning, a model that has been pre-trained on general data is provided with a small amount of labeled data so that it can learn new concepts and patterns. Our specific model outputs labels for each word in a given sentence; the sequence of labels then represents the story structure of that passage. The structure of these sequences can also be visualized as a tree. We followed a similar labeling format as the Universal Dependencies (UD) treebanks (Marneffe et al. 2021; Nivre et al. 2020), which are nested, hierarchical, and used widely in NLP tasks across many languages. Once we had labeled data for our given texts, we used several straightforward methods of analysis to explore some examples of Latin epic poetry. From the sequence of model-generated labels, we extracted and compared the labels based on their frequency and the common subsequences that occur across texts. The rest of this section describes each element in our computational pipeline in detail.

## 3.1 Data

For this study, we used a set of two Latin literary texts, which come from the Tesserae Project benchmark of intertexts (Forstall et al. 2015) between Vergil's *Aeneid* (Vergil 1987) and Lucan's *Bellum Civile* (Lucan 1987). There is extensive literary scholarship on the relationship of these two texts, whether they are seen as complementary (Mayer 1981; Rossi 2000) or as antithetical (Marti 1975; Martindale 1993; Quint 1982; Tartari Chersoni 1979; Von Albrecht 1970) to each other, which guided our usage of them here as an exemplar of two intertwined and yet unique texts that can be scrutinized using a story grammar approach.

The raw text files for these texts were obtained through the Perseus Digital Library (*Perseus Digital Library* 2025). These are the main texts used throughout this work, however an additional text was used in developing the few-shot learning hand-labeled dataset. The text used for that small set was from Vergil's *Eclogues* (Virgil 1987). We used the first 30 lines of the second poem as our few-shot dataset. We chose to use text from Vergil for our few-shot dataset, so that the model could see how these labels are applied to a text that is similar to our actual text without providing the model with the text we would be testing on.

We manually labeled the 30 examples for the few-shot dataset by following the rules of the proposed grammar and an understanding of the Latin texts. The examples were hand-labeled by a graduate student with about four years of experience working with the language as data for computational analysis and one year of formal instruction in the language. The labeled data was then verified by an expert with advanced understanding of Latin language and literature. By evaluating the purpose of each word, such as syntax and part-of-speech, in each poetic line, we were able to then determine how it fit under the hierarchy of our proposed story grammar.

All of these texts were preprocessed and cleaned by dividing the texts into individual poetic lines — which we refer to as passages throughout our work. Each passage individually has the story grammar process applied, as if it is its own small narrative. After we had divided our texts into passages, we did minimal cleaning of the data, such as removing excess characters like line numbers and book headings, but leaving any punctuation and capitalization to retain as much of the literary context as possible in the texts and to follow a similar process that future researchers might take when utilizing an LLM and our method. The two main texts, *Aeneid* and *Bellum Civile* were then labeled through the few-shot prompting method.

### 3.2 Proposed Grammar

In the study of reading, psychologists suggest that story structure and elements are the key to a reader's comprehension of a given text (Taylor 1983; Zahoor and Janjua 2013). This literary tool is particularly useful when reading prose or epic poetry, where one finds coherent narratives. We adapt a modified story grammar that has been considered before for quantitative narrative analysis (Franzosi 2010). For our use case, we have eliminated the use of exact vocabulary that maps to the terminals in the grammar, and we changed the start symbol from 'dispute' to 'narrative' to better map to our application here. Keeping 'dispute' would limit the generalizability of our grammar by limiting what vocabulary is directly linked to an exact label. We wanted to leave flexibility in our computational pipeline to allow for adapting to new literary domains during training. Our story grammar is presented here:

<narrative> ⟶ {<event>} {<document>}

<event> ⟶ {semantic triplet}

<semantic triplet> ⟶ {<subject>} {<action>} [{<object>}]

<subject> ⟶ <individual> | <set of individuals> | <institution>

<action> ⟶ <verb>[<negation>][<modality>]<circumstances>

<object> ⟶ <subject> | <physical object>

<individual> ⟶ <name> [<characteristics>]

<name> ⟶ [<first name>] [<last name>] | <name of group>

<characteristics> ⟶ [<gender>] [<occupation>] [<age>] [<work organization>] ...

<set of individuals> ⟶ <name> [<characteristics>]

<circumstances> ⟶ <time> <place> [<type>] [<reason>] [<instrument>] [<outcome>]

<instrument> ⟶ <physical object> ...

<object> ⟶ <individual> | <physical object>

This grammar is used similarly to other transformational grammars (Chomsky 1956) with <narrative> as our start symbol and all other elements as non-terminals. A narrative represents the overarching structure of the story that connects characters, events, and settings and represents their relationship to each other. We do not include any terminals in our grammar because we want to be flexible in what words or phrases can fall under each element based on the literary data we are using. If we limit the grammar to a specific vocabulary, then we remove the ability to generalize to a wide variety of literary texts, including those in different languages. Another feature of our story grammar is that order has no effect on the transformational rules, which allows for more flexibility in how these labels will be applied to a given text passage. This flexibility, in terms of word order, is particularly important in the context of Latin text. Latin typically has subject-object-verb (SOV) word order, but this is not strict. Therefore, our method has the ability to capture any word order presented in the text. The non-terminals represent the possible story element labels that are used.

### 3.2.1 Novelty of Our Labeling Structure

Here we explain the differences between our method — our proposed grammar and labeling scheme — and the existing syntactical labeling schemes that come from standard syntax parsing.

**From syntactic structure to narrative function.** Standard grammatical analysis describes the formal relationships between words, *e.g.*, a given noun functions as the subject, a given word is an adverb modifying the verb, a given phrase is a prepositional complement. Syntactic parsing is powerful, but it treats formally equivalent structures as equivalent regardless of their narrative content. An adverb is an adverb whether it specifies *when* something occurred, *where* it occurred, *why* it occurred, or *by what means*. A standard grammar does not distinguish among these functions, because it does not need to. Story grammar analysis, by contrast, treats these distinctions as primary, because they carry different information about the structure of the narrative event. An adverbial expression specifying time (denoted below as circum-time) does different narrative work than one specifying reason (circum-reason) or outcome (circum-outcome). Similarly, a subject who is a named individual (ind-name) occupies a different narrative role than one identified only by characteristics (ind-char) or membership in a collective (subj-group). The standard grammar may be identical across these cases; the narrative function is not.

**The semantic triplet as the core unit.** The fundamental unit of the proposed story grammar is the *semantic triplet*: a subject, an action, and (optionally) an object, together with the circumstances that surround the action. Every narrative event, regardless of syntactic complexity, reduces to this structure — someone does something (to something), somewhere, at some time, for some reason, with some instrument and some outcome. The label for a semantic triplet (denoted below as trip) designates each such triplet, and a single poetic line (labeled event) may contain more than one. A line describing two agents acting, or embedding one action within another, will contain multiple triplets within a single event. event and semantic triplet are therefore purely structural brackets — they specify nesting levels rather than content. When the common subsequence “event, trip, subj, subj-ind, ind-name” is reported, it should be read not as a flat list of co-equal elements but as a path *down the label tree*, from the passage level through the structural unit to the leaf that carries the analytical signal. The stylistically meaningful information always resides at the leaf.

**The full label hierarchy.** The complete taxonomy comprises approximately 28 labels organized across five levels, as shown below. The three top-level bracket labels (narrative, event, trip) are structural; all analytical content is encoded in the terminal labels beneath subj, act, and obj.

```
narrative                    ← the whole text
  └── event                  ← one passage (one poetic line)
        └── trip             ← one semantic triplet (subject–action–object)
              ├── subj                   ← who is acting
              │     ├── subj-ind         ← an individual
              │     │     ├── ind-name   ← their name
              │     │     └── ind-char   ← a characteristic (role, age, gender...)
              │     ├── subj-group       ← a collective
              │     └── subj-inst        ← an institution
              │
              ├── act                    ← what is happening
              │     ├── act-verb         ← the verb itself
              │     ├── act-neg          ← negation (non, nec, nullus...)
              │     ├── act-mod          ← modality (nefas, licet, placet...)
              │     └── act-circum       ← circumstances of the action
              │           ├── circum-place
              │           ├── circum-time
              │           ├── circum-reason
              │           ├── circum-instrument
              │           ├── circum-outcome
              │           ├── circum-type
              │           └── circum-num
              │
              └── obj                    ← what is acted upon
                    ├── obj-subj         ← when the object is itself an agent
                    ├── obj-physobj      ← a physical thing
                    └── obj-char         ← a characteristic of the object
```

**What the labels make visible.** Consider three formulations of roughly equivalent propositional content:

- *Aeneas fled the city at dawn.*
- *Driven by fate, the Trojan hero departed before sunrise.*
- *The walls of Ilium were abandoned by their last defender as morning broke.*

A dependency parser assigns each a different syntactic analysis: they differ in voice, subject type, and surface structure. Story grammar analysis assigns all three the same underlying label sequence: **named individual → departure action → place → time**. The framework abstracts away from surface variation in order to describe what the passage is doing at the level of narrative function.

This abstraction makes aggregate comparisons meaningful in ways that syntactic or lexical analysis cannot replicate. The finding that Vergil employs circum-place significantly more often than Lucan at the opening of passages is not a claim about vocabulary or topic distribution. It is a claim about narrative method: that Vergil habitually orients the reader spatially before action commences, anchoring each episode in a setting before the agents within it are set in motion. This is consistent with the fundamentally itinerant structure of the *Aeneid*, organized as it is around a sequence of arrivals and departures. Lucan, by contrast, more frequently opens passages with agents or actions, suppressing spatial framing in favor of abrupt, action-driven transitions — a tendency scholars have long associated with his rejection of conventional epic teleology. The elevated ind-name frequency in Lucan (1.047 against Vergil’s 0.641) is similarly legible in relation to established scholarship: Lucan’s obsessive cataloguing of proper names — in the army lists, the geographical excursuses, the roll-calls of the dead — is a noted feature of the *Bellum Civile*, and the labeling recovers it systematically. In both cases, the method derives its conclusions from structural patterns alone, without reference to the semantic content of the passages in question. The capacity to do so is, we argue, the principal analytical advantage of the story grammar approach over methods that operate at the lexical or embedding level.

### 3.3 Model

As previously mentioned, the model used for the story grammar labeling task is the most current GPT model as of this writing, GPT-5 (OpenAI 2025). We selected this model because of its state-of-the-art capabilities and ease with which it can be used to prototype an application of our approach for humanists. We aim to demonstrate how this tool can be utilized by scholars to aid their research across a range of applications in literary studies. When considering using these types of models in literary analysis, the instinct is to provide the model with the text and ask direct inquiries to answer a research question. For example, if someone wanted to study the themes of Lucan’s *Bellum Civile*, they might provide an LLM or chatbot with a file containing the original text and prompt the model with a question to aid their analysis: ”Can you give me a thematic analysis of Lucan’s *Bellum Civile*?” The hope is that the model will return a coherent, and reliable, response that details the different themes across the entire text. While this approach has merit, we believe the additional numerical analysis that our method provides allows a scholar to have more control over the analysis performed with the assistance of the model, rather than allowing the model to have all of the control in making (what are frequently unoriginal) choices by directly asking for a literary analysis of a given text.

Because we have created this new labeling scheme, there is no existing data available for this specific task (except for the data that we manually labeled ourselves specifically for the experiments in this paper, which is described below). This means that we do not have a ground-truth for the labels we are applying to our data. Therefore, we performed a qualitative analysis to verify the efficacy of LLM output. As an example, consider this small excerpt from the 19th century novel, *Alice’s Adventures in Wonderland*:

> ”I can’t go back to yesterday because I was a different person then.”

When we take this passage and apply our labeling structure – inspired by the Universal Dependencies treebank – the output looks like this:

> (event (trip (subj (subj-ind **I**)) (act (act-neg **can’t**) (act-verb **go**) (act-circum (circum-place **back**) (circum-time **to**) (circum-time **yesterday**) (circum-reas **because**)))) (trip (subj (subj-ind **I**)) (act (act-verb **was**) (act-circum (circum-num **a**)) (obj (obj-char **different**) (obj-subj **person**) (obj-char **then**)))))

To highlight the hierarchical structure of the story grammar and how this can be applied, the tree structure for the same output can be seen below:

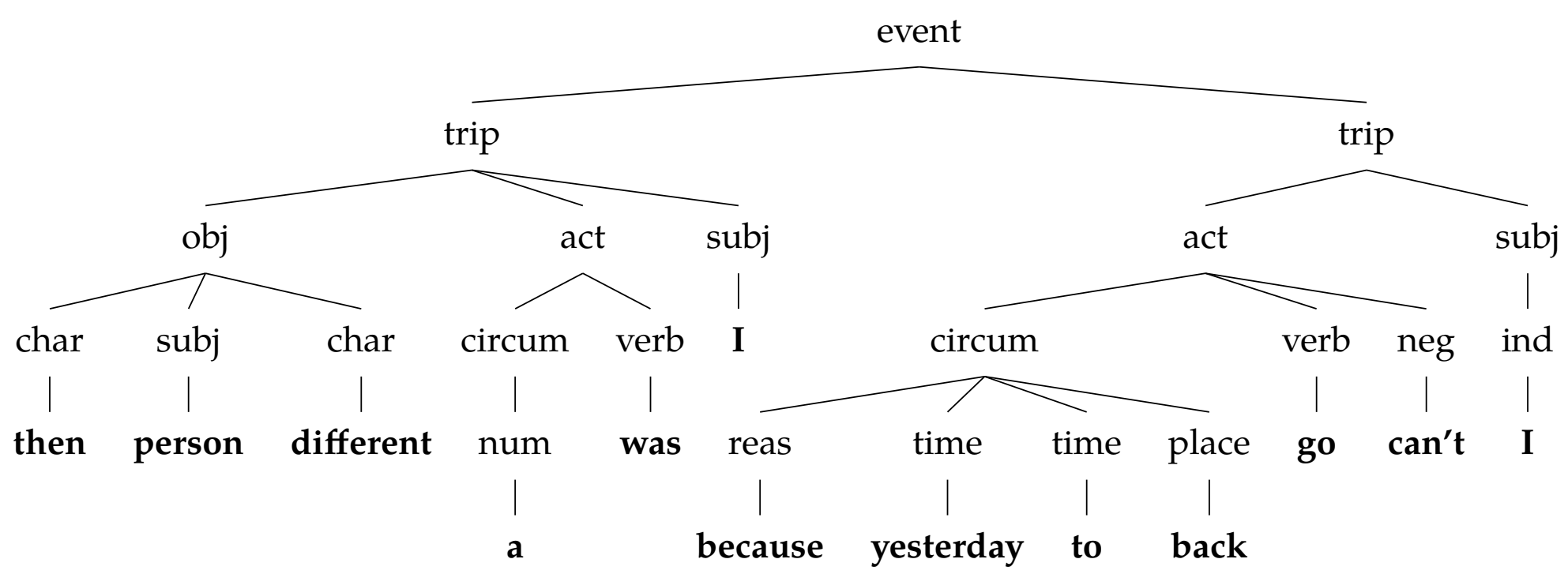

During our semi-automatic labeling process, we did not provide the model with exact specifics about when a label should be applied to a word. After performing a qualitative analysis of a sampling of the results, it is clear that the model developed its own understanding of the labels we provided and has applied them successfully to the input texts.

We cannot know for sure what data the GPT model was trained on, but we can safely assume that the proportion of classical Latin data to other languages, especially English, is tiny. As noted earlier, we used our small set of hand-labeled data to perform few-shot learning (Wang et al. 2020a). Thus, our additional few-shot learning examples would provide the model with the correct context for our given task and the language we are working with while still utilizing the extensive training the model has already gone through. This process is described in more detail below.

### 3.3.1 Few-Shot Learning Process

We provided our small set of 30 hand-labeled examples for the few-shot learning setup to the model via prompts. We used a straightforward prompting scheme. We provided the model with a few different pieces of information: the general task description, the labeling scheme as a grammar, a list of all the possible labels to use, a file containing 30 labeled examples, a template for output formatting, and a short description of the input text file that needed to be labeled. After this initial prompting, we used some intermediary prompting to fine-tune the output labeled data. An example of such additional prompting was an adjustment to the overuse of the 'obj-char' label. After our initial prompt, a review of the output data revealed that the 'obj-char' label was being overused by the model as a catchall for sometimes large portions of passages. With additional prompting, we were able to refine the models' usage of the 'obj-char' label which improved the model's labeling results. This process was fairly simple and very time efficient, so it may easily be applied to pipelines for other humanities research.

## 4. Initial Results & Analysis

Once we have passed the texts and prompt to the GPT-5 model, we end up with labeled output as described in the previous section. With these labeled passages, we can perform a stylometric analysis that is easily interpretable by technologists and humanists alike. For each of the texts, we counted how many times each of the labels was used throughout. We can use these counts as a metric to understand how the structure of a story might unfold in each text by noting the patterns in the frequency of story element usage within one text. We can further analyze these counts by looking at how these story element frequencies vary across different texts and different authors. In our experiments we compared the label counts from the *Aeneid* to the label counts from *Bellum Civile*.

The second form of analysis we conducted was to find all of the most common — or most frequently used — subsequences that occurs throughout each text. A subsequence is any consecutive sequence of story grammar labels that correspond to a given passage. These subsequences represent common story grammar label patterns that are utilized by the author of the text. We found analyzing these subsequences to be another rich source of patterns that still remains accessible to non-technologists. We performed this analysis between our main two texts of epic poetry, *Aeneid* and *Bellum Civile*, and a Latin prose text from Livy, *Ab urbe condita* (Livy 1884–1911). This demonstrates how these story grammar labels can be applied to a different type of text and how we propose researchers can compare them.

In Table 1, we list the various label counts for each text as it was labeled through our experiment pipeline. By analyzing the label counts for each text, we can better understand what story structure elements are used more or less often throughout a text. We can use these numbers to then understand aspects of Lucan and Vergil's style as it pertains to specific story elements. These counts are not normalized for the length of each text, but we note that each text had very similar lengths for each of their individual books. Therefore, the numbers we have for the *Aeneid* are slightly higher in general due to the fact that it has 12 books while *Bellum Civile* only has 10.

We also evaluated each of the texts across their individual books to see if there are any trends that appear within the texts themselves. For the *Aeneid*, we observed that the labels were generally quite consistent across each of the books. The numbers remained relatively equal in frequency depending on the relative length of each given book. We did notice one small trend in which the use of circumstance-instrument did increase through the text with each book. This increase was not large, but we did note it as an interesting area for further investigation.

Within *Bellum Civile*, we identified more interesting trends that occurred across the progression of the text. We noticed that there was a significant use of act-verb in Book 9 compared to their other books, paired with a significant decrease in usage of act-verb in Book 10. We also noted that this trend is similar with the circumstance-place label as well. Another small trend that we discovered is the usage of subject-group and its companion group-name in Book 1 mostly. Other than these small trends, similarly to *Aeneid*, the labels were generally consistent across the books. We believe that this maps with the consistency of style that the two authors, Lucan and Vergil, have across their own writing, and showcases that there are similarities between these two deeply intertextual works.

| Label | *Aeneid* | *Bellum Civile* |
|---|---|---|
| act | 7840 | 6822 |
| act-circum | 4081 | 3420 |
| act-mod* | - | 82 |
| act-neg* | - | 1181 |
| act-verb | 6228 | 5498 |
| circum-instr | 309 | 348 |
| circum-num | 416 | 116 |
| circum-out | 384 | 500 |
| circum-place | 4950 | 3791 |
| circum-reas | 208 | 256 |
| circum-time | 807 | 695 |
| circum-type | 149 | 170 |
| event | 9897 | 8060 |
| group-name | - | 34 |
| ind-char | 1153 | 212 |
| ind-name | 6350 | 8436 |
| obj | 9837 | 8035 |
| obj-char | 33556 | 25943 |
| obj-subj | 4562 | 4057 |
| subj | 5922 | 6313 |
| subj-group | - | 34 |
| subj-ind | 7503 | 8648 |
| trip | 9897 | 8060 |
| unk | 6126 | - |

**Table 1:** Story grammar label counts for the *Aeneid* and for *Bellum Civile*. These label counts provide an additional analysis of what story elements are used more or less frequently by each author and can be used to further analyze each author's style. For reference, *Bellum Civile* has 8,060 poetic lines and *Aeneid* has 9,897 poetic lines.

When comparing the two full texts, we used the total label counts for each book that can be seen in Table 1. It is important to note that these numbers represent the number of times each of these labels appeared across the model's labeling of each text. These numbers are not normalized, so to make a direct comparison between the texts, we must also consider the differing lengths of each text, which will contribute to some of the differences between the counts listed in the table. In this paper, we provide both the raw label counts in Table 1, as well as the relative frequency of each label in Table 2. We chose to display the raw numbers to preserve the full literary context that might be removed by normalizing the numbers. However, when we highlight trends here we provide context for the values by discussing them in relation to the text length as to provide the reader with an understanding of how this number might be affected by the length of the text. Therefore, we have also provided the frequency that each label is used in a text by the number of poetic lines in each text. Thus, a frequency in Table 2 represents the number of times a label is used per poetic line in the text.

Looking at the raw numbers in Table 1, it may seem that *Aeneid* has a higher usage of the act-verb label, but rather *Bellum Civile* uses this label slightly more with a frequency of 0.682 compared to 0.629 from *Aeneid*. Some other surprising or interesting trends we noted from these label counts is the difference in the frequencies for the labels that relate to subjects and individuals: subj-ind, ind-char, ind-name. We can see that *Bellum Civile* uses the subj-ind label with a frequency of 1.073 – which means that several passages in the text used this label multiple times within one passage. This can be compared to the usage of this label in the *Aeneid* which had a frequency of 0.758, which is lower than the frequency in *Bellum Civile*. When we go one step lower in the story grammar hierarchy to the ind-name and ind-char labels, we see a split in usage between the texts. *Bellum Civile* uses the ind-name label at a much higher frequency of 1.047 compared to the frequency in the *Aeneid* which was 0.641. We believe that this is a particularly interesting trend in the stylometric differences between the two texts, especially due to the army catalogues, in Books 1 and 3, and geographical lists in *Bellum Civile*. The opposite trend is true for the ind-char label which has *Aeneid* with the higher frequency at 0.116 compared to the frequency for *Bellum Civile* at 0.026.

| Label | *Aeneid* | *Bellum Civile* |
|---|---|---|
| act | 0.792 | 0.846 |
| act-circum | 0.412 | 0.424 |
| act-mod* | 0.000 | 0.010 |
| act-neg* | 0.000 | 0.147 |
| act-verb | 0.629 | 0.682 |
| circum-instr | 0.031 | 0.043 |
| circum-num | 0.042 | 0.014 |
| circum-out | 0.039 | 0.062 |
| circum-place | 0.500 | 0.470 |
| circum-reas | 0.021 | 0.032 |
| circum-time | 0.082 | 0.086 |
| circum-type | 0.015 | 0.021 |
| event | 1.000 | 1.000 |
| group-name | 0.000 | 0.004 |
| ind-char | 0.116 | 0.026 |
| ind-name | 0.642 | 1.047 |
| obj | 0.994 | 0.997 |
| obj-char | 3.391 | 3.219 |
| obj-subj | 0.461 | 0.503 |
| subj | 0.598 | 0.783 |
| subj-group | 0.000 | 0.004 |
| subj-ind | 0.758 | 1.073 |
| trip | 1.000 | 1.000 |
| unk | 0.619 | 0.000 |

**Table 2:** Story grammar label frequencies for the *Aeneid* and for *Bellum Civile*, where a frequency is the number of times a label is used divided by the number of poetic lines in the text. These label frequencies provide an additional analysis of what story elements are used more or less frequently by each author and can be used to further analyze each author's style. For reference, *Bellum Civile* has 8,060 poetic lines and *Aeneid* has 9,897 poetic lines.

One area in need of improvement that we discovered through our computational pipeline is the difference in which the model used the act-mod and act-negation labels for each text. As seen in Table 1, no label count is listed for the *Aeneid* for both act-mod and act-neg. After further investigation of the output of labeled data from the model, we found that for different "negative" Latin words such as *non*, *nec*, *nullo*, *nullus*, and *nulla*, the model treated each of the words differently even across the two texts. For *non*, we found that the model consistently labeled this word as act-neg for *Bellum Civile* while in *Aeneid* it was labeled as obj-char. We saw this exact same trend for the word *nec* as well. For *nullo*, *nulla*, and *nullus*, the model was slightly less consistent in labeling them as obj-char, occasionally using ind-name when the word was capitalized in the given passage. This highlights the importance of preprocessing and normalizing the text before it is passed to a model. We chose not to preprocess the text in our experiments to model the easiest way through the pipeline to mimic the most likely way this process might be used by researchers in the future. This shows an avenue for further refinement in our prompting technique to mitigate this small error in future experiments.

| Subsequence | *Aeneid* | *Bellum Civile* |
|---|---|---|
| event, trip, subj, subj-ind, ind-name | 4769 | 6094 |
| obj-char, obj-char | 22481 | 16942 |
| obj-char, obj-char, obj-char | 13734 | 9862 |
| event, trip, act, act-circum, circum-place | 349 | 73 |
| event, trip, subj, subj-ind, ind-name, obj, obj-char | 749 | 719 |

**Table 3:** A sample of the most frequently used story grammar subsequences in both the *Aeneid* and *Bellum Civile*. These subsequences represent common story structure patterns that each author used more or less frequently throughout their respective text, which provides a new lens for analysis.

Even with this difference in labeling schemes across texts, we will still note that *Bellum Civile* does contain more negation than *Aeneid*, which was discovered through manual investigation. This shows why the model may have struggled with labeling negation in the *Aeneid*, with fewer examples and the nuance of which these words can be used in Latin text. For the act-mod label, we believe the model struggled with the meaning of this label and where it fit into the hierarchy of our grammar. With only 82 labels used in *Bellum Civile*, the model was consistent with the words it labeled (*e.g.*, *nefas*, *licet*, *placet*), although it still did not follow the exact definition we had envisioned. This is another area in need of refinement for future experiments. However, this highlights the flexibility of our method that does not restrict the model to using the labels only for specific words in a given dictionary, but rather allows the model to easily adapt and change through our prompting scheme.

In Table 3, we present some of the most interesting results among the most common subsequences for each text. For each text, we pulled the top 25 most commonly used subsequences across the entire text. The subsequences that are used the most frequently across a text represent a common story structure pattern that the author used while writing the text; therefore, we can analyze these patterns to better understand the style and narrative structure used by a given author and a given text. From these common patterns, we noted the differences and similarities between the two texts. These examples show some of the common patterns that both Lucan and Vergil use in their respective texts; thus, we view them as a representation of the intertextuality between these two texts.

For both texts, we found that the most common start of a passage — which can be noted by the labels "event, trip" at the beginning of the subsequence — was "event, trip, subj, subj-ind, ind-name". This highlights some similarity in the style of both Lucan and Vergil in the usage of names and subjects at the start of passages. We can also see that two other common patterns that start a passage were shared across the texts. This includes "event, trip, subj, subj-ind, ind-name, obj, obj-char," which highlights the SOV word order that is common in Latin text. We also found that "event, trip, act, act-circum, circum-place" was much more common in the *Aeneid* than in *Bellum Civile*. This demonstrates Vergil's tendency to start a passage by naming a place, which Lucan does not share, at least in *Bellum Civile*.

| Label | Count | Frequency |
|---|---|---|
| act | 559 | 0.995 |
| act-circum | 543 | 0.966 |
| act-mod* | - | 0.000 |
| act-neg* | 256 | 0.456 |
| act-verb | 2408 | 4.285 |
| circum-instr | 268 | 0.477 |
| circum-num | 64 | 0.114 |
| circum-out | 208 | 0.370 |
| circum-place | 989 | 1.760 |
| circum-reas | 161 | 0.286 |
| circum-time | 292 | 0.520 |
| circum-type | - | 0.000 |
| event | 562 | 1.000 |
| group-name | - | 0.000 |
| ind-char | 2483 | 4.418 |
| ind-name | 324 | 0.577 |
| obj | 545 | 0.970 |
| obj-char | - | 0.000 |
| obj-subj | 9993 | 17.781 |
| subj | 495 | 0.881 |
| subj-group | - | 0.000 |
| subj-ind | 495 | 0.881 |
| trip | 562 | 1.000 |
| unk | - | 0.000 |

**Table 4:** Story grammar label counts and frequencies for Book 1 of Livy's *Ab urbe condita*, where a frequency is the number of times a label is used divided by the number of poetic lines in the text. Both of these values provide an additional analysis of what story elements are used more or less frequently by an author and can be used to further analyze an author's style.

Because both of these texts are epic poetry, we wanted to study how these different analyses would compare to prose text. We used the same computational pipeline on a digitized version of Book 1 of Livy's *Ab urbe condita* (Livy 1884–1911). The passage was passed to the GPT-5 model with the same exact prompt and few-shot data as our other experiments. This produced labeled text using our story grammar and we did the same analysis of counting the labels and finding the most commmon subsequences. The resulting label counts and frequencies can be seen in Table 4. The main difference we discovered between the results for epic poetry text and prose text was the difficulties of working with longer, more complex passages. The length of each of the sentences in the Livy text are considerably longer than passages in either the *Aeneid* or *Bellum Civile*. Because we used the same few-shot data from our other experiments, which was poetic text, the model struggled to apply the information learned from those examples to this new style of text.

We believe that this underscores the most important element in applying our method in future work. If a small few-shot dataset with the same style of text can be created for the prompting of the model, the model should be able to adapt the story grammar to a new set of data. This also shows that in the prompting for the few-shot learning method, the dataset is indeed internalized by the model to understand how the grammar should be applied to the input text. One interesting thing that the model did was overuse the obj-subj label across almost every input passage with a very high frequency of 17.781, meaning that each sentence in the text would have, on average, 17 words that map to the object-subject story element. As discussed previously, when refining our prompt with the original experiments, we had to steer the model away from overusing the obj-char label. This does introduce a promising area of further investigation into the inherent differences between epic poetry and prose that might lead the model into preferring the obj-subj or obj-char label over the other.

## 5. Case Study: Semantic Matching

For stylistics (Craig 2004), the goal is to understand and compare the style of literary texts by genre and author. Computational methods (Rybicki et al. 2016) for this task utilize the information and features of embeddings of the input text. Our method can provide an additional feature set for style — similarly to how we compared Lucan to Vergil using our common subsequences and label counts. These features can be added directly into existing computational pipelines for this task. The same is true for authorship attribution (Stamatatos n.d.; Tyo et al. 2022), which possesses similarities to computational stylistics. Similarly, allusion detection (Bamman and Crane 2008; Evans 1988; Hinds 1998) and semantic matching (Chandrasekaran and Mago 2021) share a relation in their task formation for machine learning. To demonstrate the potential of our story grammar method for improving and changing the approach to these various machine learning tasks, we provide a case study in applying the story grammar labels to a semantic matching algorithm.

The semantic matching task can be defined formally. A pair of texts, $T_1$ and $T_2$, can be divided into smaller text passages $t_i$ where $1 < i <= N_p$ and $N_p$ is the number of passages that make up a given text. Given a model $M$, the set of text passages $\tau_1$ and $\tau_2$ for our pair of texts $T_1$ and $T_2$, respectively, can be processed to produce a set of features $f_1$ and $f_2$. These feature sets can then be used to compare the sets of text passages. Each text passage in $\tau_1$ is compared to each text passage in $\tau_2$ using their respective feature sets $f_1$ and $f_2$ and a given metric $\mu$ which results in a set of scores $S$. For a given threshold value $\Theta$, each set of text pairs that have a score $s_i \leq \Theta$ are considered a semantic match.

For each of the input texts — the same we used in our previous experiments — we parsed out just the labels from our labeled output, so that we would have sequences of story grammar labels. We then treated each of these sequences as a representation for its respective input passage. The method then compares these pairs of texts using a metric $\mu$, which for our method is a weighted cosine similarity score. Cosine similarity measures the distance between vectors in space by finding the cosine angle between them. The algorithm for calculating the cosine similarity where $A$ and $B$ are vectors is:

$$cosine(a, b) = \frac{A \cdot B}{||A||\,||B||}$$

Our method processes the sequences of labels and considers each one to be their own tree. We use a different methodology to calculate the cosine similarity between two labeled passages. We calculate three different similarity scores and then combine them together using weights to get our final similarity score for a pair of passages. We calculate a cosine similarity for the nodes, edges, and paths individually. For our application, nodes represent each individual label in a sequence, edges represent the connections between parent and child labels in a sequence — such as the subject to subject-individual relationship or the act to act-verb relationship — and finally paths represent an extension of edges with a specific length $n$. For our work, we chose a length of $n = 3$ for our paths. This gives us a final similarity score calculation that looks like the following, where $w_{nodes}$, $w_{edges}$, and $w_{paths}$ are the respective weights for each part of the overall score:

$$similarity = w_{nodes} \cdot cosine_{nodes} + w_{edges} \cdot cosine_{edges} + w_{paths} \cdot cosine_{paths}$$

From this comparison method, the output is a matrix of calculated weighted cosine similarity scores, where one dimension of the matrix represents the passages from the source text and the other dimension represents the passages from the target text. Thus, the matrix shows the full comparison of every passage in each text being compared with every passage in the opposite text. We compare our results to the existing Tesserae benchmark on our input texts (Forstall et al. 2015).

Here is an example of what our output will look like outside of the matrix of possible parallels:

**Source:** *Aeneid* 4.628

**Text:** *Litora litoribus contraria, fluctibus undas*
**Translation:** shore against shore and the opposing seas [*trans. Williams, Theodore, C, 1910.*]
**Labeled:** (event (trip (act (act-circum (circum-place ***Litora***) (circum-place ***litoribus***))) (obj (obj-char ***contraria***) (obj-char **fluctibus**) (obj-char ***undas***))))

**Target:** *Bellum Civile* 1.6

**Text:** *In commune nefas, infestisque obvia signis*
**Translation:** into shared guilt, and against hostile standards
**Labeled:** (event (trip (act (act-mod ***nefas***) (act-circum (circum-place ***in***) (circum-place ***commune***))) (obj (obj-char ***infestisque***) (obj-char ***obvia***) (obj-char ***signis***))))

**Score:** 0.975345

This example of a highly ranked match in our results between Book 1, line 6 from *Bellum Civile* and Book 4, line 628 from *Aeneid* shows how our method uses the story grammar labeling scheme to provide additional information for each of the passages. The GPT-5 model, using our prompts and refinements, produced these labels for both passages and after applying the weighted cosine similarity metric on the label sequence — excluding any of the actual Latin text — we can see that this pair has a score of approximately 0.975. Theoretically, the more similar two passages are, the closer their weighted cosine similarity score should be to 1. This passage pair has a very high score and this was actually the highest ranking parallel for this particular line in *Bellum Civile*. Furthermore, this passage pair is also one of the intertexts provided in the Tesserae benchmark that is a verified parallel by literary commentators (Forstall et al. 2015). Our method was able to recover a number of other commentator parallels with little lexical correspondence in the Tesserae benchmark. Overall, applying the story grammar labels in this downstream task of semantic matching recovered 501 of the total 3,410 possible parallels in the top 10% of results we produced.

## 6. Discussion

While our results are less abstract than working with typical machine learning outputs, there still remains a measure of interpretation that must be performed to process the output. We aimed to have our method recall and align with traditional analyses of these two epic poems, but also produce novel ideas and highlight new areas of interest in both texts. One clear alignment with standard literary analyses of the two poems emerges in the distribution of passage-initial location labels (circum-place). We found that the *Aeneid* more frequently opens passages with explicit place references than the *Bellum Civile*. This accords with the fundamentally itinerant structure of Vergil's epic, which is organized around a sequence of arrivals and departures — Troy, Carthage, Sicily, Cumae, and Latium among many others — and often uses spatial framing to orient the reader at the outset of an episode.

Lucan's poem, by contrast, while spanning a wide geographical range (from Rome and Spain to Thessaly, Africa, and Egypt), more often suppresses extended scene-setting in favor of abrupt narrative transitions and action-driven openings. This stylistic difference reflects Lucan's broader rejection of epic teleology and traditional narrative orientation, even in episodes with strong spatial focus, such as the Erichtho scene. The differing frequencies of circum-place thus capture a meaningful contrast in narrative technique between the two poets and illustrate how our method can recover established literary insights while grounding them in systematic textual patterns.

Using our story grammar method allows humanists to identify large-scale patterns of narrative organization in a text without requiring exhaustive close reading of every passage. More importantly, the method makes visible recurring habits in a poet's compositional practice — how action is introduced and resolved, how scenes are framed or omitted, how conflict is staged, and how pacing is managed across a work. Because these patterns are defined at the level of narrative function rather than lexical repetition, the method can recover meaningful similarities and contrasts even when poets employ different vocabularies or stylistic registers.

In this respect, story grammar analysis aligns closely with expert literary reading: it abstracts away from individual word choice in order to model how semantic roles and narrative structures operate across a text as a whole. The results therefore speak directly to familiar concerns of literary criticism — setting, agency, action, and thematic emphasis — rather than to opaque statistical features or embedding spaces. At the same time, the method produces outputs that are systematic, reproducible, and interpretable, allowing scholars to move fluidly between quantitative patterns and qualitative analysis. By making narrative structure legible at scale while remaining grounded in humanistic categories, this approach supports both independent literary interpretation and richer collaboration between humanists and technologists.

The immediate results of our method themselves are a direct way to analyze an input text; however, the output is easily transformed into a feature set that can be used in a multitude of other machine learning tasks for literary analysis. These could include: stylistics, authorship attribution, allusion detection, and semantic matching. In other work, these tasks are treated as a standard NLP task using the text directly and creating embeddings as a unit for comparison for analyzing the texts. However, such a strategy is limited in its interpretability, unlike our method which has an output that is easily understood and readable from the initial output. Because our method results in a feature set that is different from the input and output of other computational methods, it complements them and is compatible for combining various methods for ensemble learning. Due to the granularity of our labels, by word, and our processing, by passage, this pairs well with methods that have a different level of granularity. Overall, combining the strengths of other methods with the advantages of our story grammar labels can lead to a richer representation of input literary text.

## 7. Conclusion

Distant reading methods frequently rely on vector-space embeddings and abstract statistics to represent downstream literary analysis tasks. Our method introduces a new perspective for computational literary analysis by utilizing story grammars. The use of story grammars in the mental model of readers for literary comprehension inspired our use of story element labels as a new approach for performing computational literary analysis. We used a GPT-5 model to label our Latin epic poetry texts and used this output to explore the similarities and differences between the two texts and their authors. We highlight the various ways this output can be used for analysis from a simple label count to searching common subsequences and even to downstream literary analysis tasks built for machine learning such as semantic matching. We showcased the utility of this new approach and propose that the interpretability of the results and its ease of use makes this approach accessible for both humanists and technologists alike. We suggest that the qualities of our method, which require some understanding of both machine learning and scholarly reading, should foster a richer collaboration for further exploration of computational literary analysis.

## 8. Data Availability

Data can be found here: `https://anonymous.4open.science/r/Modeling-Story-Grammars-E86B`

## 9. Software Availability

Software can be found here: `https://anonymous.4open.science/r/Modeling-Story-Grammars-E86B`

under review

## 10. Author Contributions

**Abigail Swenor:** Methodology, Data curation, Software, Writing – original draft

**John James:** Validation, Writing - review & editing

**Neil Coffee:** Validation, Writing – review & editing

**Walter Scheirer:** Conceptualization, Supervision, Writing – review & editing